\documentclass[12pt]{article}
\usepackage[preprint,nonatbib]{neurips_2024}
\usepackage[round]{natbib}  
\usepackage[T1]{fontenc}

\usepackage{ltablex}    
\usepackage{booktabs}   
\usepackage{amssymb}    
\usepackage{hyperref}   
\keepXColumns         

\usepackage{pgfplots}
\usepackage{graphicx}
\usepackage{amsmath}
\usepackage{amsfonts}
\usepackage{hyperref}
\usepackage{longtable}  
\usepackage{adjustbox}   
\usepackage{geometry}
\usepackage{algorithm}
\usepackage{svg}
\usepackage{algpseudocode}
\usepackage{amssymb}
\usepackage{tikz}
\usepackage{cleveref} 
\usetikzlibrary{arrows.meta, shapes, positioning, quotes}

\usepackage{xcolor}

\usepackage[most]{tcolorbox} 
\tcbuselibrary{listings, breakable} 
\usepackage{listings}
\usepackage{booktabs}

\title{AIDE: AI-Driven Exploration in the Space of Code}
\author{
  Zhengyao Jiang\footnotemark[1] \footnotemark[3]\\
  Weco AI\\
  \And
  Dominik Schmidt\footnotemark[1] \footnotemark[2]\\
  Runway ML\\
  \And
  Dhruv Srikanth\\
  Weco AI\\
  \And
  Dixing Xu\\
  Weco AI\\
  \And
  Ian Kaplan\\
  Weco AI\\
  \And
  Deniss Jacenko\\
  Weco AI\\
  \And
  Yuxiang Wu\footnotemark[1] \footnotemark[3]\\
  Weco AI
}
\date{}

\begin{document}
 
\maketitle

\footnotetext[1]{Equal contribution.}
\footnotetext[2]{Work done while at Weco AI.}
\footnotetext[3]{Corresponding to \texttt{zhengyao@weco.ai} and \texttt{yuxiang@weco.ai}.}

\begin{abstract}
Machine learning, the foundation of modern artificial intelligence, has driven innovations that have fundamentally transformed the world.
Yet, behind advancements lies a complex and often tedious process requiring labor and compute intensive iteration and experimentation. Engineers and scientists developing machine learning models spend much of their time on trial-and-error tasks instead of conceptualizing innovative solutions or research hypotheses.
To address this challenge, we introduce AI-Driven Exploration (AIDE), a machine learning engineering agent powered by large language models (LLMs). 
AIDE frames machine learning engineering as a code optimization problem,
and formulates trial-and-error as a tree search in the space of potential solutions.
By strategically reusing and refining promising solutions, AIDE effectively trades computational resources for enhanced performance, achieving state-of-the-art results on multiple machine learning engineering benchmarks, including our Kaggle evaluations, OpenAI's MLE-Bench and METR's RE-Bench.
The implementation of AIDE is publicly available at \url{https://github.com/WecoAI/aideml}.
\end{abstract}

\section{Introduction}
Machine learning engineering supports many modern AI achievements, from basic regression on tabular data to the recent surge in large generative models.
However, building a high-performance machine learning model is always time consuming.
Due to the inherent stochasticity of both the data and the optimization process, engineers and scientists rely heavily on trial-and-error.
Researchers have long sought to automate these iterative processes, 
leading to advancements in fields like AutoML 
\citep{Feurer2015,autosklearn,h2oautoml,tpot,autokeras2023,autokeras2019,autoweka-jmlr,autoweka-kdd,autogluon}, 
Neural Architecture Search 
\citep{Zoph2017,Pham2018,Liu2019,Real2019,nas-lit-review}, 
and hyperparameter optimization 
\citep{Falkner2018,hpo-lit-review}.
These methods typically require a predefined search space of configurations, such as hyperparameters and network architectures, within which the algorithm explores potential solutions~\citep{nas-lit-review,hpo-lit-review,white2023neural}.
Defining this space often requires significant domain expertise.
Furthermore, search algorithms for hyperparameter tuning are often somewhat brute force compared to human experts, resulting in lower compute efficiency and a risk of overfitting to the validation set.

The emergence of advanced coding capabilities in large language models (LLMs)~\citep{gpt4-tech-report,swebench,first-claude35sonnet-system-card,team2024gemini,jain2025livecodebench,o1-system-card,o3-system-card} has introduced an exciting new possibility: searching directly within the space of code rather than the space of predefined configurations.
Code-space optimization offers greater flexibility and leverages the extensive domain-specific knowledge inherent in LLMs, effectively narrowing the search to more promising solutions and thus boosting sample efficiency.
This gives it the potential to address compute-bound tasks like deep learning or, presumably, even optimizing LLMs themselves.

Here, we introduce AI-Driven Exploration (AIDE), a LLM-powered agent\footnote{In this paper, “Agent” may refer either to the algorithm built on top of LLMs or to the entire system with LLMs included, depending on the context.} that automates the trial-and-error process of machine learning engineering. Unlike the ReACT~\citep{Yao2023} style agent, which appends historical observations to the LLM’s context and relies on the model's capabilities to solve a monolithic optimization problem, AIDE organizes all historical solutions in a tree structure. It then asks the LLM to propose improvements based on individual tree nodes. A hard-coded tree-search algorithm accumulates these incremental improvements, guided by automated evaluations.

We benchmarked AIDE on a set of Kaggle tasks focusing on tabular machine learning and released these initial results together in April 2024~\citep{aide-report}.
Subsequently, OpenAI released MLE-Bench~\citep{mlebench}, further showing that AIDE can be applied to even more challenging deep learning tasks from Kaggle while achieving state-of-the-art performance.
Most notably, AIDE achieves twice the number of medals compared to a follow-up agent~\citep{openhands-platform} when both use GPT-4o; with o1-preview, the gap widens even further.
In parallel, METR assessed AIDE on AI research tasks against human experts under time constraints, showing that AIDE can outperform expert-crafted solutions in limited time windows~\citep{rebench}.
Moreover, for tasks with a robust evaluation signal like Triton Kernel optimization, AIDE’s final solution surpasses that of human experts, even when the latter had extended development time.

The first half of this paper provides a formal specification of AIDE for the research community. In the second half, we present and analyze empirical evaluations of AIDE, drawing on both our own experiments and independent benchmark results.

\section{Preliminaries}
Many general-purpose LLM agents, including ReACT \citep{Yao2023}, frame their tasks as Partially Observable Markov Decision Processes (POMDPs) \citep{POMDP}, a widely used framework in reinforcement learning.
In a POMDP, the agent tries to maximize a cumulative reward by choosing actions based on all past observations, essentially treating the entire interaction history as the state.
While this approach is flexible and unifies a range of tasks, it lacks a principled way to break down the problem when there is a clear structure available. 
Moreover, for LLM-based agents, continually appending all historical data can lead to oversized prompts and limit scalability, because the model’s context window eventually fills up. 


In this work, we adopt an alternative framework for LLM-driven iterative problem solving by modeling the task as an optimization problem: 
Let \(\mathcal{S}\) be a space of possible solutions (e.g., Python scripts), and let \(h: \mathcal{S} \rightarrow \mathbb{R}\) be a \emph{stateless} objective function (for example, validation accuracy or loss). The goal is to find an optimal solution:

\begin{equation}
  s^* = \arg\max_{s \in \mathcal{S}} h(s).
\end{equation}

Each candidate solution $s$ can be evaluated independently via an objective function $h(s)$. This perspective simplifies the problem considerably: rather than unrolling a single, long-horizon decision process , we can directly evaluate and compare solutions. It also aligns naturally with existing optimization methods, like tree search, which depend on standalone evaluations of candidate solutions.

\section{Methodology}
\label{sec:methodology}
In this section, we introduce our approach to automating machine learning engineering with AIDE.
By employing the tree search method, AIDE systematically explores solutions that optimize validation metrics, breaking down the monolithic optimization task into atomic improvement steps.
We begin by outlining the high-level optimization algorithm.
And then delve into key implementation details, such as the search policy and specialized prompts that drive the iterative generation and refinement of machine learning code.

\subsection{AI-Driven Exploration in the Space of Solutions}
\label{sec:ssts}

In AIDE, a \textbf{solution} \(s\) is the code to be optimized, with \(s_0\) denoting the empty root solution.
An \textbf{evaluator}, \(h: \mathcal{S} \to \mathbb{R}\), evaluates the code and provides a scalar score.
All discovered solutions are stored in a \textbf{solution tree}, \(T\), whose nodes correspond to scripts and edges represent an improvement attempt (e.g., \(s \to s'\) is an improvement of
\(s\)).
A \textbf{search policy}, \(\pi(T)\), selects which solution
\(s \in T\) will serve as the base solution to be improved.
To keep language model prompts concise while being aware of the historical attempts,
a \textbf{summarization operator}, \(\Sigma(T)\), extracts relevant information
from the tree, such as the high level idea of each improvement attempt and its corresponding  performance metrics.
Finally, a \textbf{coding
operator}, \(f\bigl(s,\Sigma(T)\bigr)\), proposes new scripts by drafting an
initial version from \(s_0\), fixing bugs, or refining a promising solution based
on the summarized context.

With these components in place, AIDE
can systematically explore the code solution space, as shown in Algorithm~\ref{alg:aide}.

\begin{algorithm}[t]
\caption{AI-Driven Exploration (AIDE)}
\label{alg:aide}
\begin{algorithmic}[1]
\State \textbf{Initialize}: solution tree $T_0 \leftarrow \varnothing$
\State \textbf{Initialize}: base solution $s \leftarrow s_0$

\For{$n = 1, 2, ..., N$}
    \State $s_n \leftarrow f\bigl(s,\, \Sigma(T_{n-1})\bigr)$ 
    \Comment{Propose a new solution}
    \State $v_n \leftarrow h(s_n)$ 
    \Comment{Evaluate the solution}

    \State $T_n \leftarrow T_{n-1} \cup \{\text{node }(s_n, v_n), \text{edge }(s \to s_n)\}$
    \Comment{Record node and its score}

    \State $s \leftarrow \pi(T_n)$ 
    \Comment{Select the next base node}
\EndFor

\State \textbf{return } $\operatorname{argmax}_{s' \in \{s_0, ..., s_N\}} h(s')$  \Comment{Best solution found}
\end{algorithmic}
\end{algorithm}

\subsection{AIDE for Machine Learning}
\label{sec:aide}
Here we present more implementation details of AIDE for machine learning engineering, providing a concrete instantiation of the core components from Section~\ref{sec:ssts}. In particular, we build upon the following design elements:

\paragraph{Search Policy (\(\pi\)).} 
In AIDE, the search policy \(\pi\) (~\cref{alg:aide}, line 7) follows a simple hard-coded rule, determining whether to draft, debug, or improve based on an existing solution. Specifically, it selects:
\begin{itemize}
    \item \textit{Drafting} if we have not yet reached the desired number of initial solutions.
    \item \textit{Debugging} if a buggy node remains within a certain debug depth.
    \item \textit{Improving} otherwise, typically targeting the best (non-buggy) solution.
\end{itemize}
This policy imposes practical heuristics, such as 1) first exploring a set of diverse initial solutions and continuously improving the best one, and 2) constraining the number of debug attempts for a broken solution.

\paragraph{Coding Operator (\(f\)).} 
 The coding operator has three main entry points, each with its own specialized prompts:
\begin{itemize}
    \item \emph{Drafting}, which is invoked when we need a completely new solution from scratch. It prompts an LLM to outline a brief plan for a model (e.g., specifying a particular network architecture or feature-engineering idea), then emits a single-file Python program implementing that plan.
    \item \emph{Debugging}, which focuses on repairing buggy solutions. By inspecting error logs and execution traces, it attempts to rectify issues in the code like broken imports, incorrect tensor dimensions, or other coding errors while preserving the overall approach.
    \item \emph{Improving}, which is called when a valid, non-buggy solution already exists but could benefit from data preprocessing, architectural or optimization modifications. Here, the LLM proposes exactly one “atomic” change, such as switching optimizers or adding a regularization technique, so that its effect on performance is directly measurable.
\end{itemize}
Combining these three operations keeps the solution tree structured and ensures that each new node arises from a well-defined modification of a parent node.

\paragraph{Summarization Operator (\(\Sigma(T)\)).}
Despite the flexibility to generate arbitrarily large numbers of solutions, we avoid saturating the LLM’s prompt by applying a context summarization operator, \(\Sigma(T)\). Instead of appending all historical logs, \(\Sigma(T)\) selectively extracts:
\begin{itemize}
    \item Performance metrics (e.g., accuracy, AUC-ROC, test set loss).
    \item Hyperparameter settings if a solution involves a hyperparameter sweep.
    \item Relevant hints for debugging (e.g., misaligned array shapes in tracebacks).
\end{itemize}
A concise summary is crucial to maintaining a stateless perspective: each code revision stands on its own, but \(\Sigma(T)\) uses prior information to guide subsequent proposals. This design offers much of the benefit of incremental reasoning without exploding the prompt size.

\paragraph{Data Preview in Coding Prompts.}
In addition to dynamic updates from \(\Sigma(T)\), AIDE for machine learning includes a small static “data preview” in each prompt, giving the LLM basic knowledge of dataset size or feature layouts. In practice, we store relevant metadata (e.g., number of rows, column names, or data splits) in the workspace and insert it into the coding operator’s prompt. Although not a complete EDA pipeline, this lightweight approach helps AIDE guide key code decisions. These decisions include selecting a validation split or scaling hyperparameters, without repeatedly including extensive dataset context.

\begin{figure}[t]
    \centering
    \
    \definecolor{brandBgLight}{HTML}{F0F0F0}    
    \definecolor{brandBg}{HTML}{F0EADA}         
    \definecolor{brandAccent}{HTML}{FD4578}     
    \definecolor{brandBlack}{HTML}{0D0F18}      
    \definecolor{brandBuggy}{HTML}{4A4A4A}      
    
    \begin{tikzpicture}[
        solution/.style={
            circle,
            minimum size=0.8cm,
            draw=brandBlack,
            thick,
            font=\normalsize
        },
        root/.style={
            solution,
            fill=brandBgLight
        },
        bug/.style={
            solution,
            fill=brandBuggy,
            text=white
        },
        improvement/.style={
            solution,
            fill=brandBg,
        },
        optimal/.style={
            solution,
            fill=brandAccent,
            text=white,
            double,
            double distance=0.5pt
        },
        arrow/.style={
            -Stealth,
            thick,
            color=brandBlack
        },
        label/.style={
            fill=white,
            inner sep=1pt,
            font=\small
        }
    ]
    \node[root] (s0) at (0,0) {$s_0$};
    \node[bug] (s1) at (-3.2,-2) {$s_1$};
    \node[improvement] (s2) at (0,-2) {$s_2$};
    \node[bug] (s3) at (3.2,-2) {$s_3$};
    \node[bug] (s4) at (-4.4,-4) {$s_4$};
    \node[improvement] (s5) at (-2,-4) {$s_5$};

    \node[optimal] (s6) at (0,-4) {$s_6$};

    \node[improvement] (s7) at (3.2,-4) {$s_7$};

    \draw[arrow] (s0) -- (s1);
    \draw[arrow] (s0) -- (s2);
    \draw[arrow] (s0) -- (s3);
    \draw[arrow] (s1) -- (s4);
    \draw[arrow] (s1) -- (s5);
    \draw[arrow] (s2) -- (s6);
    \draw[arrow] (s3) -- (s7);
    \node[label] at (-2.2,-0.8) {$f\colon \text{draft}$};
    \node[label] at (0,-0.8) {$f\colon \text{draft}$};
    \node[label] at (2.2,-0.8) {$f\colon \text{draft}$};
    \node[label] at (-4,-3) {$f\colon \text{fix}$};
    \node[label] at (-2.6,-3) {$f\colon \text{fix}$};
    \node[label] at (0,-3) {$f\colon \text{improve}$};
    \node[label] at (3.2,-3) {$f\colon \text{fix}$};
    \node[root] (l1) at (5.6,0) {};
    \node[right=0.2cm of l1] {Empty Solution};
    \node[bug] (l2) at (5.6,-1) {};
    \node[right=0.2cm of l2] {Bug Detected};
    \node[improvement] (l3) at (5.6,-2) {};
    \node[right=0.2cm of l3] {Valid Solution};

    \begingroup
        \hypersetup{%
          pdfborder={0 0 0}
        }
        \node[optimal] (l4) at (5.6,-3) {\href{https://form.typeform.com/to/KjuvoA2b#hubspot_utk=xxxxx&hubspot_page_name=xxxxx&hubspot_page_url=xxxxx}{\phantom{\rule{1em}{8pt}}}}; 
    \endgroup
    \node[right=0.2cm of l4] {Optimal Solution};

    \end{tikzpicture}
    \caption{A sample solution tree \(T\) for AIDE, where each node is a Python script. 
    Arrows represent transitions proposed by the coding operator \(f\). 
    Some branches terminate in a bug, while others lead to improved or optimal solutions.}
    \label{fig:solution-tree}
\end{figure}
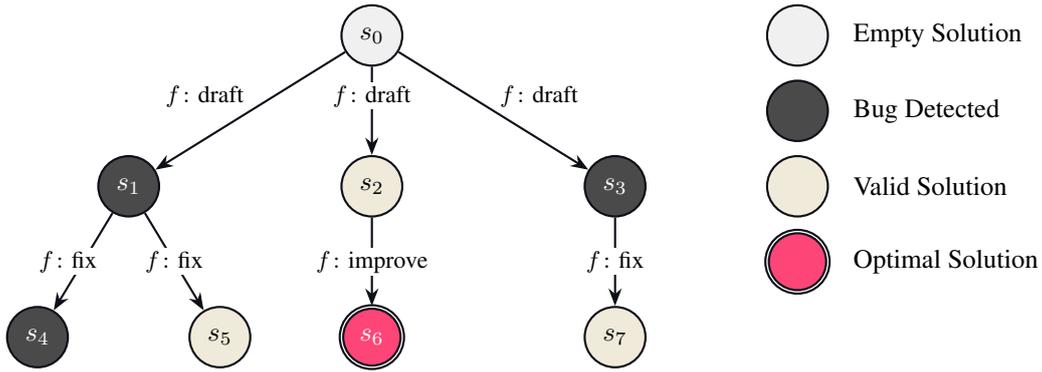

\paragraph{Putting It All Together.}
Figure~\ref{fig:solution-tree} illustrates how AIDE’s instantiation for machine learning uses (i) a search policy \(\pi\) to select which solution to refine next, (ii) a coding operator \(f\) for generating code by drafting, debugging, or improving solutions, and (iii) a summarization operator \(\Sigma(T)\) to keep the LLM prompts concise and targeted. By combining these components under a stateless optimization framework, AIDE can systematically search within the space of possible code solutions for machine learning tasks, avoiding an ever-increasing prompt history while retaining the relevant knowledge needed to achieve high performance.

\section{Evaluation}
In this section we report empirical evaluations of AIDE.
We did our own evaluation on Kaggle competitions with a focus on tabular machine learning tasks~\citep{aide-report}.
On the other hand, after the open sourcing of the AIDE in April 2024, the community has done larger scale independent evaluations showing promising results on deep learning~\citep{mlebench} and AI R\&D~\citep{rebench} tasks.
We therefore also aggregate relevant results here to provide a better understanding of the AIDE’s performance.
Readers interested in the extended evaluations are encouraged to read and cite the papers from OpenAI~\citep{mlebench} and ~\citet{rebench} respectively.

\subsection{Weco Kaggle Benchmark}

We curated a diverse set of Kaggle competitions to build Weco's internal Kaggle benchmark, called \textbf{Weco-Kaggle}, for evaluating AIDE's performance in machine learning.
This set consists of 63 competitions of varied complexity and data size, spanning domains such as tabular machine learning, image classification, and time-series prediction.
Some of these competitions require a GPU to solve.
Full details of the competitions in Weco-Kaggle are provided in Appendix~\ref{appendix:weco-kaggle-full}.
From Weco-Kaggle, we selected a subset of 16 tabular machine learning tasks with relatively lower complexity and primarily CPU-based runtime requirements.
This subset, referred to as \emph{Weco-Kaggle Lite}, is shown in Table~\ref{table:aide_internal_16_comps}.

\paragraph{Evaluation Protocol.}

We evaluate the performance of AIDE by comparing its results to that of human competitors in each Kaggle competition, and averaging across competitions. 
We follow the evaluation protocol below to evaluate AIDE's and other frameworks' performance:

\begin{enumerate}
\item  Before running the agent on a competition, we split the competition's training data into an \textit{agent train set} and a \textit{holdout test set}. This split is defined manually for each competition following similar parameters as Kaggle's official private test set (e.g. similar train-test percentages), but is not necessarily the same, since Kaggle's test set is not released publicly for most competitions. Note that our holdout test set is also distinct from the train-validation split that AIDE itself generates as part of its internal node evaluation protocol.
\item During code generation, AIDE is given access to the holdout test inputs (but not labels) and prompted to evaluate its model on this data. In particular, we prompt AIDE to generate a \texttt{submission.csv} file, analogously to how human competitors submit their competition results.
\item We define an \emph{Exceeds \% of Human} metric as $100(1 - q)$, where $q$ is the quantile of AIDE's score on the official Kaggle leaderboard. This metric represents the percentage of human competitors whose performance AIDE surpasses. Whenever possible, we use Kaggle's private leaderboard because it is less prone to overfitting by competitors; if a private leaderboard is unavailable, we default to the public leaderboard. In addition, we report the \emph{Above Median} metric, originally proposed by \cite{mlebench}, which indicates how frequently AIDE outperforms the median Kaggler performance across competitions.
\item This metric is then averaged across all competitions.
\end{enumerate}

We chose our evaluation protocol based on leaderboard-quantiles since, unlike each competition's included metric, these scores are similarly distributed between competitions, making it possible to simply average across competitions to obtain aggregated scores. Leaderboard quantiles are also more fine-grained, allowing us to evaluate, for example, the performance of a single run on a single task, unlike medal-counts \citep{mlebench} which collapse to a binary metric in this case. Finally, our scores are interpretable and useful in assessing AIDE's performance relative to humans.

\begin{table}[h]
    \centering
    \footnotesize
        \begin{tabular}{llcc}
            \hline
            \textbf{Agent} & \textbf{Model} 
            & \textbf{Exceeds \% of humans} $\uparrow$ 
            & \textbf{Above Median (\%)}  $\uparrow$ \\
            \hline
            AIDE & GPT-4 Turbo & \textbf{51.38} & \textbf{50.00} \\
            AutoML (H2O) & N/A & 35.34 & 18.75  \\
            \href{https://js.langchain.com/v0.1/docs/use_cases/autonomous_agents/auto_gpt/}{AutoGPT (Langchain)} & GPT-4 Turbo & 32.34 & 0.00 \\
            Human with ChatGPT & GPT-4 Turbo & 41.17 & 18.75 \\
            \hline
        \end{tabular}
        \vspace{1em}
    \caption{Comparing AIDE to other agent frameworks on 16 tabular machine learning tasks from Kaggle. \emph{Exceeds \% of humans} indicates the percentage of human Kaggle participants being outperformed by the agents, averaged across the competitions. \emph{Above Median (\%)} is the fraction of competitions where the score was strictly above the median of human Kaggle participants.}
    \label{table:aide_internal_results}
\end{table}



\paragraph{Baselines.}
To evaluate AIDE's effectiveness, we compare it against three baselines that illustrate different approaches to automated or assisted machine learning:
\begin{enumerate}
    \item \textbf{Conventional H2O AutoML.} 
    We select H2O, one of the leading AutoML platforms, to exemplify traditional AutoML tools. In each competition, the data is split into an 80\%/20\% train/validation set, and model selection is performed within a 600-second search window.
    \item \textbf{AutoGPT.}
    A workflow automation framework that surged in popularity in early 2024. It generates a plan and automatically executes the necessary steps to complete a task. We adapt its task descriptor to produce solutions for our competitions.
    \item \textbf{Human Assisted with ChatGPT.} 
    An increasingly common scenario involves human engineers leveraging ChatGPT to assist with coding tasks. We adopt this baseline to understand how AIDE performs relative to a human engineer directing ChatGPT to develop solutions.
\end{enumerate}
These baselines collectively provide a robust comparative foundation for evaluating AIDE against both traditional AutoML workflows and modern LLM-assisted strategies. Further details about the baselines' configuration can be found in Appendix~\ref{appendix:baseline}.

\begin{table}[ht]
\centering
 \resizebox{\linewidth}{!}{%
\begin{tabular}{l r r r c}
\hline
\textbf{Competition} & \textbf{Total Teams} & \textbf{AIDE Rank} & \textbf{Exceeds \% of Human} & \textbf{Above Median}\\
\hline
\href{http://kaggle.com/c/playground-series-s3e14}{playground-series-s3e14} & 1877 & 897 & 52.21\% & False \\
\href{http://kaggle.com/c/playground-series-s3e16}{playground-series-s3e16} & 1431 & 693 & 51.57\% & False \\
\href{http://kaggle.com/c/playground-series-s3e19}{playground-series-s3e19} & 1174 & 742 & 36.80\% & True \\
\href{http://kaggle.com/c/playground-series-s3e22}{playground-series-s3e22} & 1543 & 1142 & 25.99\% & True \\
\href{http://kaggle.com/c/playground-series-s3e24}{playground-series-s3e24} & 1910 & 655 & 65.71\% & False \\
\href{http://kaggle.com/c/playground-series-s3e25}{playground-series-s3e25} & 1633 & 948 & 41.95\% & True \\
\href{http://kaggle.com/c/tabular-playground-series-aug-2022}{tabular-playground-series-aug-2022} & 1889 & 392 & 79.25\% & False \\
\href{http://kaggle.com/c/tabular-playground-series-feb-2021}{tabular-playground-series-feb-2021} & 1434 & 559 & 61.02\% & False \\
\href{http://kaggle.com/c/tabular-playground-series-feb-2022}{tabular-playground-series-feb-2022} & 1257 & 708 & 43.68\% & True \\
\href{http://kaggle.com/c/tabular-playground-series-jan-2022}{tabular-playground-series-jan-2022} & 1592 & 886 & 44.35\% & True \\
\href{http://kaggle.com/c/tabular-playground-series-jul-2021}{tabular-playground-series-jul-2021} & 1294 & 1126 & 12.98\% & True \\
\href{http://kaggle.com/c/tmdb-box-office-prediction}{tmdb-box-office-prediction} & 1395 & 692 & 50.39\% & False \\
\href{http://kaggle.com/c/bike-sharing-demand}{bike-sharing-demand} & 3243 & 262 & 91.92\% & False \\
\href{http://kaggle.com/c/cat-in-the-dat}{cat-in-the-dat} & 1341 & 714 & 46.76\% & True \\
\href{http://kaggle.com/c/house-prices-advanced-regression-techniques}{house-prices-advanced-regression-techniques} & 4978 & 1357 & 72.74\% & False \\
\href{http://kaggle.com/c/new-york-city-taxi-fare-prediction}{new-york-city-taxi-fare-prediction} & 1485 & 819 & 44.85\% & True \\
\hline
\textbf{Average} &  &  & 51.38\% & 50.00\% \\
\hline
\end{tabular}
}
\caption{AIDE vs. human performance comparison on Weco-Kaggle Lite. The submissions were made manually in February 2024. All rankings are actual rankings on the private/public Kaggle leaderboard, assessed in February 2024.}
\label{table:aide_internal_16_comps}
\end{table}

\paragraph{AIDE's Results on Weco-Kaggle Lite.}
Table~\ref{table:aide_internal_results} compares AIDE against multiple baselines, including H2O AutoML, AutoGPT, and a human competitor utilizing ChatGPT, averaged over the 16 tabular Kaggle tasks of Weco-Kaggle Lite. AIDE achieves an \emph{Exceeds \% of humans} score of 51.38\%, outperforming half of the Kaggle participants on average, and surpasses the human median in 50\% of these tasks. By contrast, H2O AutoML and LangChain AutoGPT attain lower \emph{Exceeds \% of humans} scores (35.34\% and 32.34\%, respectively). Table~\ref{table:aide_internal_16_comps} offers a detailed breakdown for each competition, indicating that AIDE’s performance ranges from surpassing roughly 13\% of human participants (for more challenging tasks) to nearly 92\% (for tasks it handles more effectively). Across half of the competitions, AIDE ranks above the human median, underscoring its robustness in consistently delivering competitive results against a diverse set of real-world machine learning challenges.

\begin{figure}[h]
    \centering
    \includegraphics[width=0.95\linewidth, trim=0 40 0 20, clip]{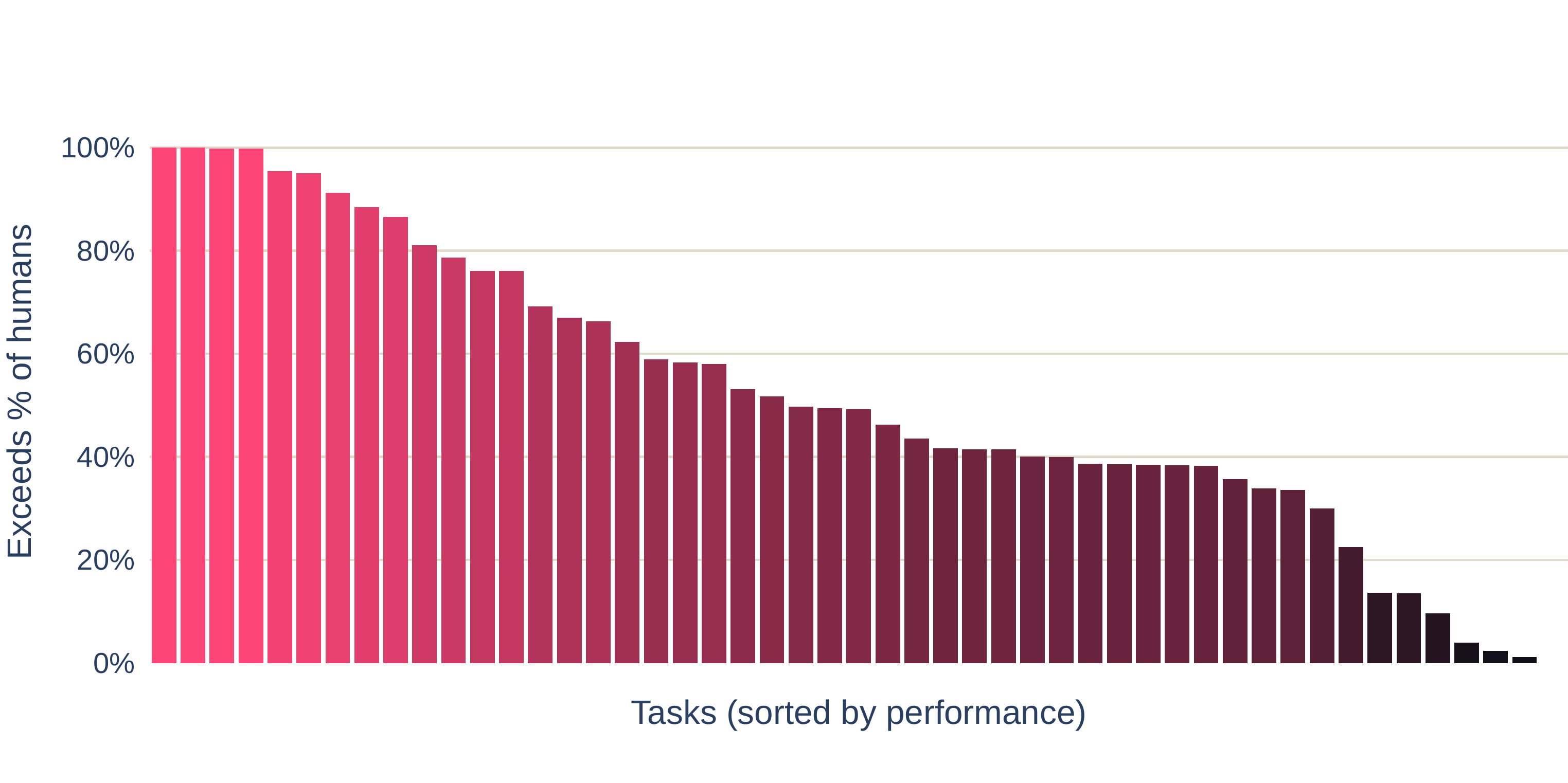}
    \caption{AIDE's performance distribution on full Weco-Kaggle benchmark. \textit{Exceeds \% of Humans} values are estimated from the leaderboard distribution.}
    \label{fig:weco-kaggle-full-aide-results}
\end{figure}

\paragraph{AIDE's Results on Full Weco-Kaggle.}
Figure~\ref{fig:weco-kaggle-full-aide-results} illustrates AIDE’s performance distribution across our extended set of Kaggle competitions, sorted by its \emph{Exceeds \% of Humans} value. Notably, AIDE achieves near-top-tier performance on several tasks, surpassing the vast majority of human participants, while on other tasks it exceeds only a small fraction. Overall, the average \emph{Exceeds \% of Humans} rate is 48.23\%, and AIDE outperforms the human median in 49.21\% of the competitions. These results underscore that AIDE can be highly competitive in certain domains, yet there remains variability in its performance depending on the dataset and task requirements.

\paragraph{Potential Limitations.}
Despite the advantages discussed above, our protocol has some limitations. First, because our test set may differ from Kaggle’s private test set, scores may not always be directly comparable, which can result in variance in percentiles.
Second, there is a risk of contamination since some of the language models used in this work may have been trained on competition-related data.
Although we found no significant correlation between agent performance and competition recency, the only way to fully ensure no data contamination would be to submit the agent's solutions to live competitions.

\subsection{AIDE in MLE-Bench}
\label{subsec:aide_mle_bench}
MLE-Bench~\citep{mlebench} is an offline evaluation framework comprising 75 real Kaggle competitions.
Here, we present the results related to AIDE reported by \citet{mlebench} and encourage readers to check and cite the original paper if they are interested in the results presented here.
In these evaluations, \textbf{AIDE} emerged as the top-performing agent framework when paired with state-of-the-art large language models. 
Other agent frameworks such as ResearchAgent from MLAB~\citep{huang2024mlagentbench} and OpenHands~\citep{openhands-platform} tended to terminate early or struggle with iterative refinement.
AIDE's optimization-centric approach led to better scalability in terms of trial-and-error interactions, therefore higher valid-submission rates and ultimately more competition medals.

Table~\ref{table:aide_results} highlights key results of AIDE compared to other agents. 
The reported \emph{Any Medal (\%)} column shows the fraction of competitions on which the agent and model combination achieved a medal (bronze, silver, or gold) in a single pass (i.e.\ pass@1). 
AIDE with o1-preview earned medals in 16.9\% of competitions, nearly four times that of the follow-up agent OpenHands.

\begin{table}[t]
    \centering
    \footnotesize
    \resizebox{\linewidth}{!}{%
        \begin{tabular}{llcccc}
            \hline
            \textbf{Agent} & \textbf{Model} 
            & \textbf{Valid Subm.\ (\%)} 
            & \textbf{Above Median (\%)} 
            & \textbf{Gold (\%)} 
            & \textbf{Any Medal (\%)} \\
            \hline
            AIDE & o1-preview 
                 & 82.8 ± 1.1 & 29.4 ± 1.3 & 9.4 ± 0.8 & 16.9 ± 1.1 \\
            AIDE & GPT-4o    
                 & 54.9 ± 1.0 & 14.4 ± 0.7 & 5.0 ± 0.4 & 8.7 ± 0.5 \\
            AIDE & Llama 3.1 
                 & 27.3 ± 2.6 & 6.7 ± 1.4  & 1.7 ± 0.7 & 3.0 ± 1.0 \\
            AIDE & Claude 3.5
                 & 51.1 ± 3.3 & 12.9 ± 2.2 & 4.4 ± 1.4 & 7.6 ± 1.8 \\
            \hline
            MLAB & GPT-4o    
                 & 44.3 ± 2.6 & 1.9 ± 0.7  & 0.8 ± 0.5 & 0.8 ± 0.5 \\
            OpenHands & GPT-4o
                 & 52.0 ± 3.3 & 7.1 ± 1.7  & 2.7 ± 1.1 & 4.4 ± 1.4 \\
            \hline
        \end{tabular}
    }
    \caption{Full MLE-Bench results (pass@1) reported by~\citep{mlebench} comparing AIDE to other agent frameworks. \emph{Valid Subm.\ (\%)} is the fraction of all competitions (not just those with a submission) where the submission passed validity checks. \emph{Above Median (\%)} is the fraction of competitions where the score was strictly above the median of human Kaggle participants. \emph{Any Medal (\%)} is the fraction awarded a bronze, silver, or gold medal (the primary success metric). Each experiment is repeated with 3 seeds, except for AIDE+o1-preview and AIDE+GPT-4o, which use 16 and 36 seeds respectively. Scores represent the mean ± one standard error of the mean.}
    \label{table:aide_results}
\end{table}
\vspace{0.2em}
\noindent

\begin{figure}[t]
    \centering
    \includegraphics[width=0.9\linewidth, trim=0 10 0 0, clip]{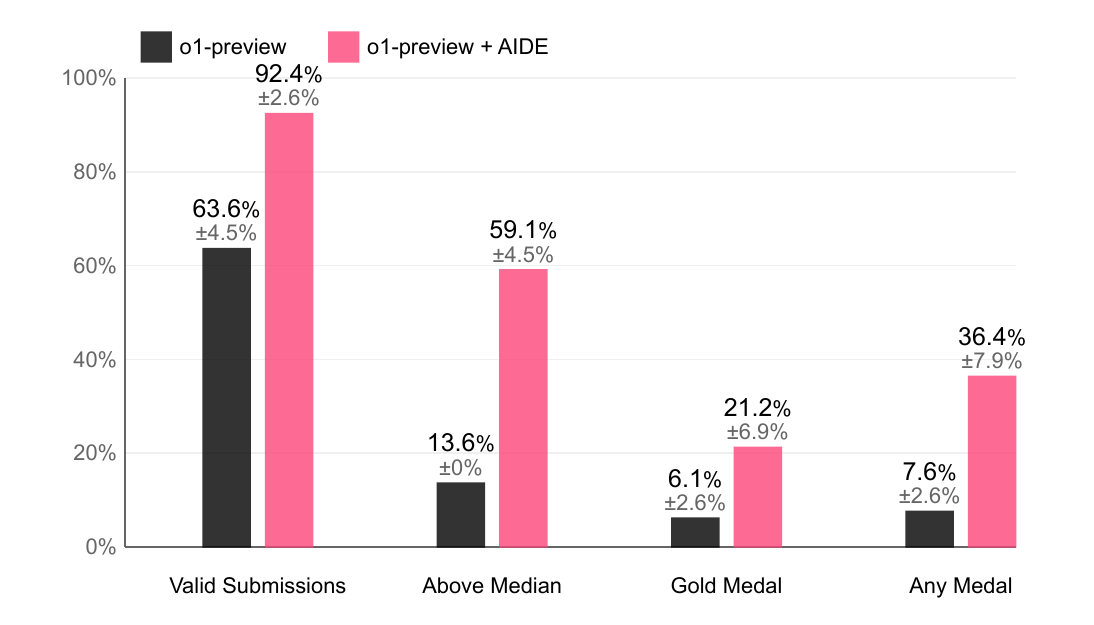}
    \caption{Performance of o1-preview with and without AIDE on MLE-bench Lite (complexity=low) set.}
    \label{fig:mle-bench-lite-with-vs-without-aide}
\end{figure}

\paragraph{AIDE’s Key Advantages.} 
By explicitly implementing a \emph{solution tree search} strategy, AIDE keeps node-level code concise and focuses each language model call on a localized problem (e.g.\ debugging only the most promising script). 
This design helps avoid oversized prompts, preserves a clear performance record for each node, and repeatedly refines partial solutions over the entire 24-hour timeframe.
Consequently, AIDE systematically addresses coding bugs and suboptimal hyperparameters rather than abandoning failed solutions. 
As shown in Table~\ref{table:aide_results}, these iterative improvements translate into higher medal acquisition rates in comparison to generic agents.

Moreover, when given additional attempts per competition (increasing $k$ in pass@\textit{k}), AIDE significantly increases its success rate; 
for instance, GPT-4o and o1-preview nearly double their medals from pass@1 to pass@6 \citep{mlebench}. 
These observations underscore the \emph{specialized} nature of AIDE, 
which often outperforms other agents through persistent, Kaggle-style iteration, 
highlighting the efficacy of a competition-targeted design in real-world ML tasks.

The impact of AIDE becomes particularly evident when comparing performance on the MLE-bench Lite subset, as shown in Figure~\ref{fig:mle-bench-lite-with-vs-without-aide}. Using o1-preview with AIDE significantly improved performance across all metrics compared to using o1-preview alone. The valid submission rate increased from 63.6\% $\pm$ 4.5\% to 92.4\% $\pm$ 2.6\%, demonstrating AIDE's effectiveness in guiding the model through the submission process. More importantly, the fraction of solutions scoring above the median human performance increased dramatically from 13.6\% to 59.1\% $\pm$ 4.5\%, and both medal-related metrics showed substantial improvements: the gold medal achievement rate more than tripled from 6.1\% $\pm$ 2.6\% to 21.2\% $\pm$ 6.9\%, while the overall medal achievement rate increased nearly fivefold from 7.6\% $\pm$ 2.6\% to 36.4\% $\pm$ 7.9\%. These improvements are statistically significant ($p < 0.01$ for all metrics, two-tailed t-test). The dramatic performance gains across all metrics demonstrate that AIDE's iterative optimization approach substantially enhances the model's problem-solving capabilities, enabling more reliable and higher-quality solutions through systematic refinement.

\subsection{AIDE in RE-Bench}
\label{subsec:aide_rebench}
While AIDE is designed for building machine learning pipelines, METR applied it to much more challenging AI R\&D tasks by formulating these tasks into optimization tasks.
The tasks range from optimizing a Triton Kernel to finetuning GPT-2 for QA.
Surprisingly, AIDE performs quite well on these tasks, and is even comparable with the top human AI scientists from Google DeepMind, Google, Anthropic,
OpenAI, FAR Labs, Redwood Research, University of California Berkeley, Carnegie Mellon University, Stanford University, or Massachusetts Institute of Technology~\citep{rebench}.

\begin{figure}[h]
    \centering
    \includegraphics[width=0.7\linewidth]{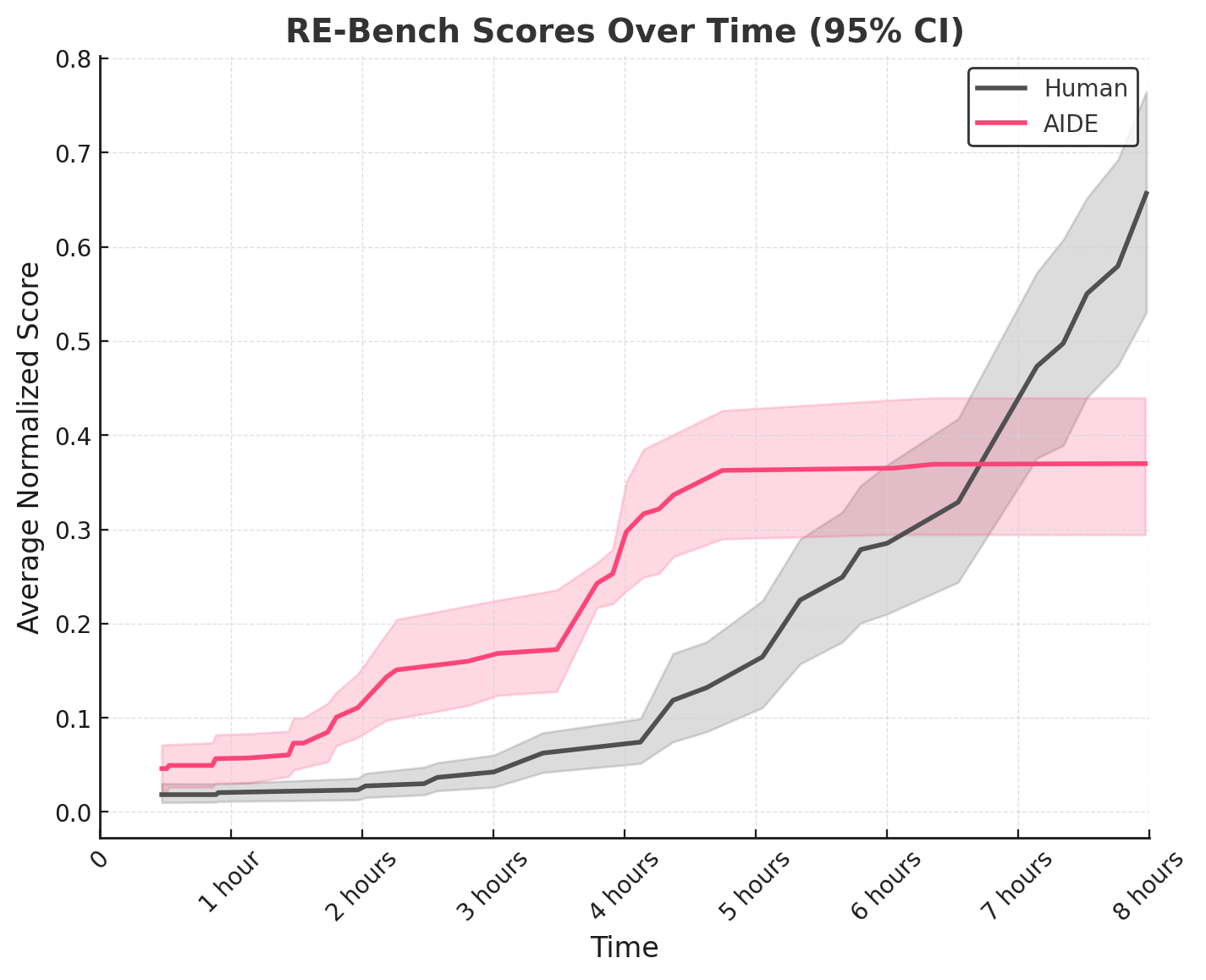}
    \caption{Average score achieved by AIDE+o1-preview and top human scientists on 7 AI R\&D tasks, as report by \citet{rebench}. AIDE managed to surpass human scientists within six hours by enabling faster experiment iterations. However, human scientists eventually caught up, as AIDE adopts a simple greedy policy that may lead to local optima on challenging R\&D tasks.}
    \label{fig:rebench}
\end{figure}

Figure~\ref{fig:rebench} illustrates AIDE's average performance over time across the seven RE-Bench environments.
Since LLMs can implement solutions much faster, allowing for more iteration cycles, AIDE managed to outperform humans within the six-hour time limit.
Notably, the agent exceeded human performance in \emph{Optimize a Kernel}, discovering a custom Triton-based solution faster than any of the nine human experts did within 64 hours.
However, AIDE fell short in environments that required handling larger codebases or where a single improvement involved multiple steps of interaction. For example, in \emph{Agent for Rust CodeContests}, AIDE was prone to repeating local patches instead of discovering new strategies.

\section{Related Work}
\subsection{LLM Agents}
Recent advances in large language models have spurred the development of agents that combine natural language reasoning with task execution. \emph{General-purpose} agents such as ReAct \citep{Yao2023} and HuggingGPT \citep{Shen2023} interleave planning with action selection to perform tasks ranging from information retrieval to multi-modal processing. In contrast, \emph{specialized} agents like Voyager \citep{voyager} and AlphaCode \citep{alphacode} are tailored to specific domains such as embodied reasoning and competitive code generation. These systems integrate execution feedback into the LLM's reasoning process, enabling iterative refinement of candidate solutions.

\subsection{Automated Machine Learning}
Automated Machine Learning (AutoML) aims to eliminate manual intervention in model selection, hyperparameter tuning, and pipeline configuration. Early frameworks such as Auto-WEKA \citep{autoweka-kdd} and TPOT \citep{tpot} employed Bayesian optimization and genetic programming, respectively, to search over predefined model spaces. Later systems like Auto-Sklearn \citep{autosklearn} and AutoGluon \citep{autogluon} have leveraged meta-learning and ensemble techniques to further improve performance. Despite their success, many conventional AutoML methods rely on static search spaces and lack the dynamic adaptability required for more complex problem settings.

\subsection{Neural Architecture Search}
Neural Architecture Search (NAS) focuses on automatically designing neural network topologies. Initial methods based on reinforcement learning \citep{Zoph2017} and evolutionary strategies \citep{Real2019} demonstrated that automated search could yield competitive architectures. Differentiable approaches such as DARTS \citep{Liu2019} have reduced the computational cost by enabling gradient-based optimization over a relaxed search space. However, NAS still faces challenges in computational expense and search space design. 
AIDE, on the other hand, avoids such problems above with code space search and efficient design exploration powered by LLMs.

\section{Conclusion}
In conclusion, we have presented AI-Driven Exploration (AIDE), an LLM Agent for machine learning engineering. By systematically drafting, debugging, and refining solutions, AIDE achieves superior performance on Kaggle tasks as well as on more research-oriented benchmarks. While developed for tabular machine learning tasks, third-party experiments show that this approach can generalize to challenges such as neural architecture search, Triton Kernel optimization, and other AI R\&D tasks. We believe AIDE represents a promising step toward the future of automated ML engineering, offering a principled way to combine iterative LLM prompting with a tree-based exploration of code solutions.

\bibliographystyle{plainnat}
\bibliography{reference}

\newpage
\appendix

\section{Baseline Specifications}
\label{appendix:baseline}

\begin{table}[h]
    \caption{Baseline hyperparameters.}
    \centering
    \begin{tabular}{l l}
    \toprule
    \multicolumn{2}{c}{\textbf{AutoGPT}} \\
    \midrule
    \textbf{Parameter}       & \textbf{Value}                  \\
    \midrule
    \texttt{agent}           & \texttt{LangChain AutoGPT}           \\
    \texttt{model}           & \texttt{gpt-4-0125-preview}                 \\
    \texttt{seed}            & 1                                  \\
    \texttt{max\_runtime}    & 600                                \\
    \midrule
    \multicolumn{2}{c}{\textbf{Human with ChatGPT}} \\
    \midrule
    \textbf{Parameter}       & \textbf{Value}                  \\
    \midrule
    \texttt{model}           & \texttt{gpt-4-0125-preview}                 \\
    \bottomrule
    \end{tabular}
    \label{tab:baseline_hyperparameters}
\end{table}

\subsection{H2O AutoML Baseline}
\label{sec:baseline_h2o}

The machine learning algorithm selection process of H2O AutoML \cite{H2OAutoML20} proceeds as follows. First, it searches over a set of six algorithms:
\begin{enumerate}
    \item Distributed Random Forest (DRF) and Extremely Randomized Trees (XRT)
    \item Generalized Linear Model (GLM) with regularization
    \item XGBoost
    \item H2O Gradient Boosting Machines
    \item Fully connected multi-layer artificial neural network (DeepLearning)
    \item Stacked Ensembles (including an ensemble of all base models and ensembles using subsets of the base models)
\end{enumerate}

It then performs a random search over a predefined grid of hyperparameter combinations, avoiding the computational expense of an exhaustive grid search. After training individual models, H2O AutoML creates stacked ensembles by combining the predictions of the best-performing models from each algorithm. This ensemble method leverages the strengths of multiple models to improve overall performance. All trained models, including individual models and ensembles, are evaluated using cross-validation and ranked based on performance metrics such as accuracy, AUC, or RMSE, depending on the problem type. The configurations are shown in Table~\ref{tab:baseline_hyperparameters}.

\subsection{AutoGPT Baseline}
\label{app:autogpt}

We use the LangChain implementation of \href{https://github.com/Significant-Gravitas/Auto-GPT}{AutoGPT}, which includes LangChain primitives such as \texttt{PromptTemplates}, \texttt{VectorStores}, and \texttt{Embeddings}. Inspired by \cite{huang2024mlagentbench}, we introduce a task descriptor for AutoGPT in each competition to provide a basic task planner and minimize human intervention. The task descriptor includes information retrieved from the Kaggle page, such as the dataset description, file details (\texttt{train.csv}, \texttt{test.csv}, \texttt{sample\_submission.csv}), evaluation metrics, submission file format, and a sample training script. An example task descriptor is shown in Figure~\ref{fig:task_descriptor.txt}.

We also provide the agent with tools to read and write files, list directories, and run Python REPL evaluations. The agent reads the task descriptor with predefined goals, as shown below:

\begin{tcolorbox}[colback=lightgray!10, colframe=black, width=\textwidth, arc=2mm, boxrule=0.5mm, title=Goal Prompt]
\begin{lstlisting}[basicstyle=\small\ttfamily, breaklines=true]
Go through task_descriptor.txt to understand the task and evaluation method. Iterate over different models or feature selections to get a better performance based on evaluation method.
You can use following steps as reference:
1. Select a model and fill in the provided python snippet.
2. Train the model and Make predictions on data from test.csv and Prepare submission.csv by executing the script wiith python repl tool.
3. Save the script into local disk such as model_{model_name}.py
            
Here are some rules to follow:
1. Never try to change the train.csv and test.csv.
2. Never output graphs or figures.
3. Do Not change the capitalization of the column name
4. Do Not read train.csv and test.csv directly.
\end{lstlisting}
\end{tcolorbox}

\begin{figure}[hp]
    \centering
    \caption{The task descriptor for bike-sharing-demand}
    \begin{tcolorbox}[colback=lightgray!10, colframe=black, width=\textwidth, arc=2mm, boxrule=0.5mm, title=Task Descriptor Prompt]
    \begin{lstlisting}[basicstyle=\small\ttfamily, breaklines=true]
Dataset Description
See, fork, and run a random forest benchmark model through Kaggle Scripts

You are provided hourly rental data spanning two years. For this competition, the training set is comprised of the first 19 days of each month, while the test set is the 20th to the end of the month. You must predict the total count of bikes rented during each hour covered by the test set, using only information available prior to the rental period.

Data Fields
datetime - hourly date + timestamp  
season -  1 = spring, 2 = summer, 3 = fall, 4 = winter 
holiday - whether the day is considered a holiday
workingday - whether the day is neither a weekend nor holiday
weather - 1: Clear, Few clouds, Partly cloudy, Partly cloudy
2: Mist + Cloudy, Mist + Broken clouds, Mist + Few clouds, Mist
3: Light Snow, Light Rain + Thunderstorm + Scattered clouds, Light Rain + Scattered clouds
4: Heavy Rain + Ice Pallets + Thunderstorm + Mist, Snow + Fog 
temp - temperature in Celsius
atemp - "feels like" temperature in Celsius
humidity - relative humidity
windspeed - wind speed
casual - number of non-registered user rentals initiated
registered - number of registered user rentals initiated
count - number of total rentals

Files
train.csv - the training dat
Columns: datetime,season,holiday,workingday,weather,temp,atemp,humidity,windspeed,casual,registered,count
test.csv - the same format as train.csv, but without the target value; your task is to predict the value for each of these targets.
Columns: datetime,season,holiday,workingday,weather,temp,atemp,humidity,windspeed
sample_submission.csv - a sample submission file in the correct format.
Columns: datetime,count

Evaluation
Submissions are evaluated one the Root Mean Squared Logarithmic Error (RMSLE).

Submission Format
Your submission file must have a header and should be structured in the following format:
datetime,count
2011-01-20 00:00:00,0
2011-01-20 01:00:00,0
2011-01-20 02:00:00,0
...
...

# Load and prepare the data
train_data = pd.read_csv('train.csv')
test_data = pd.read_csv('test.csv')

# please continue with the python script for training

# Prepare submission file
submission = pd.DataFrame({'datetime': test_data['datetime'], 'count': test_predictions})
submission.to_csv('submission.csv', index=False)

    \end{lstlisting}
    \end{tcolorbox}
    \label{fig:task_descriptor.txt}
\end{figure}

\subsection{ChatGPT with Human Assistance}
\label{app:human_with_chatgpt}

A human operator is tasked with solving a Kaggle competition using only the information provided in the overview and data tabs, which include the available dataset. The operator is permitted to utilize the ChatGPT web interface. The LLM is set to  \texttt{gpt-4-0125-preview} in comparison with AutoGPT. Due to limitations in ChatGPT's capabilities, such as the potential for generating hallucinated results and occasionally using outdated packages, iterative interactions are required. The human operator will continue to issue instructions until a valid submission is produced. Upon completion, the operator submits the results to Kaggle, where the submission is ranked against the competition leaderboard.

\section{Analysis of AIDE}

\subsection{Code Complexity Growth}

\begin{figure}[h]
    \centering
    \includegraphics[width=0.8\linewidth]{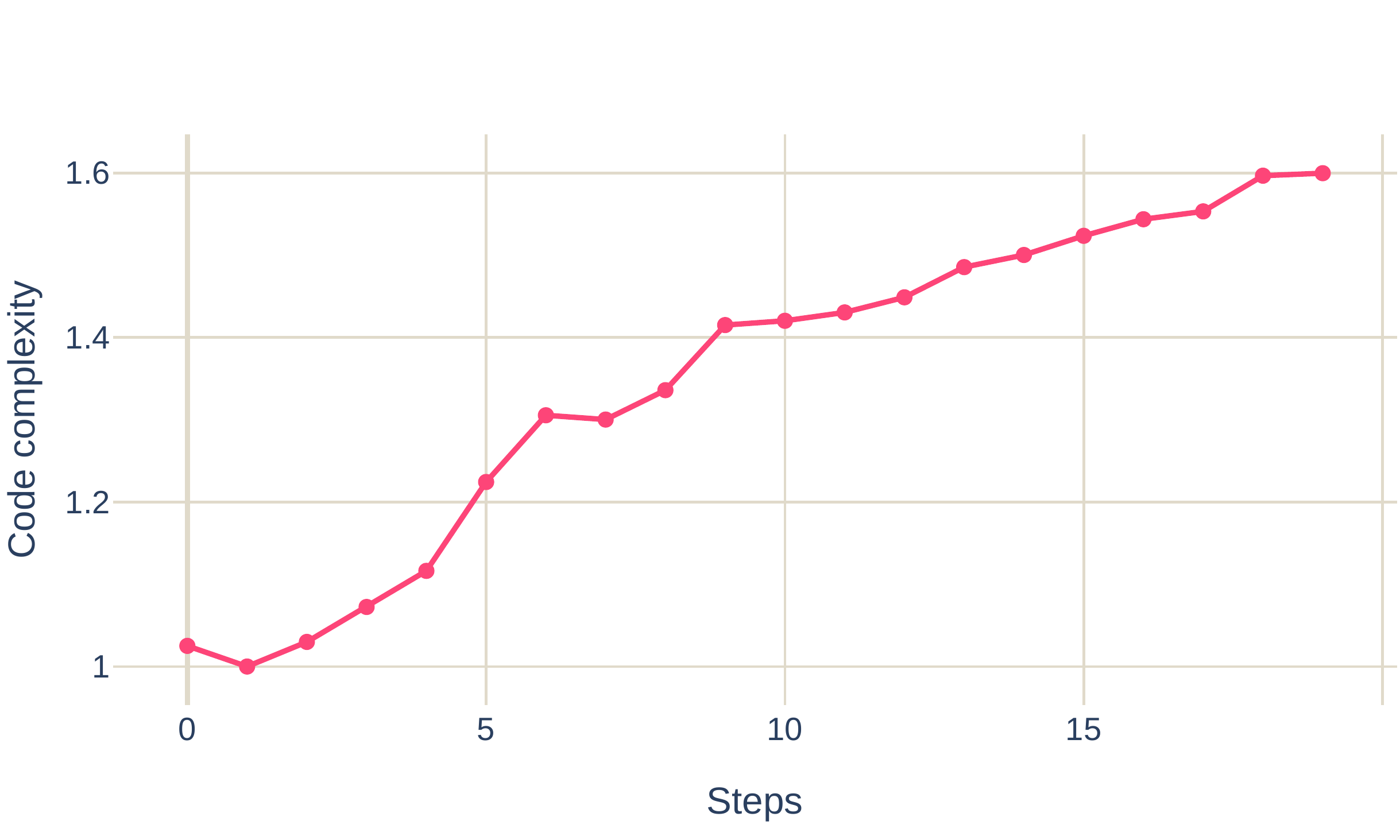}
    \caption{Aggregated code complexity (LOC, LLOC, Volume, N1, MI) of the code generated by AIDE (GPT-4 Turbo) with respect to number of steps.}
    \label{fig:aggr-cc}
\end{figure}

\par In Figure~\ref{fig:aggr-cc}, we observe that the aggregated code complexity (combining LOC, LLOC, Volume, N1, and MI) exhibits an overall increasing trend as the number of iterative steps grows. Initially, there is a slight dip in complexity, but after the first step, the metrics begin a generally steady rise. This suggests that as AIDE (GPT-4 Turbo) produces successive iterations of code, the solutions tend to become more elaborate, with additional lines of code and logical structures contributing to higher values for traditional software complexity measures. The progressive increase implies that, over multiple generation steps, the model accumulates more intricate functionality—potentially reflecting deeper problem-solving processes or additional features—leading to an increasingly complex codebase by the final iteration.

\subsection{Cost Analysis}

\begin{figure}[h]
    \centering
    \includegraphics[width=\linewidth, trim=0 20 0 0]{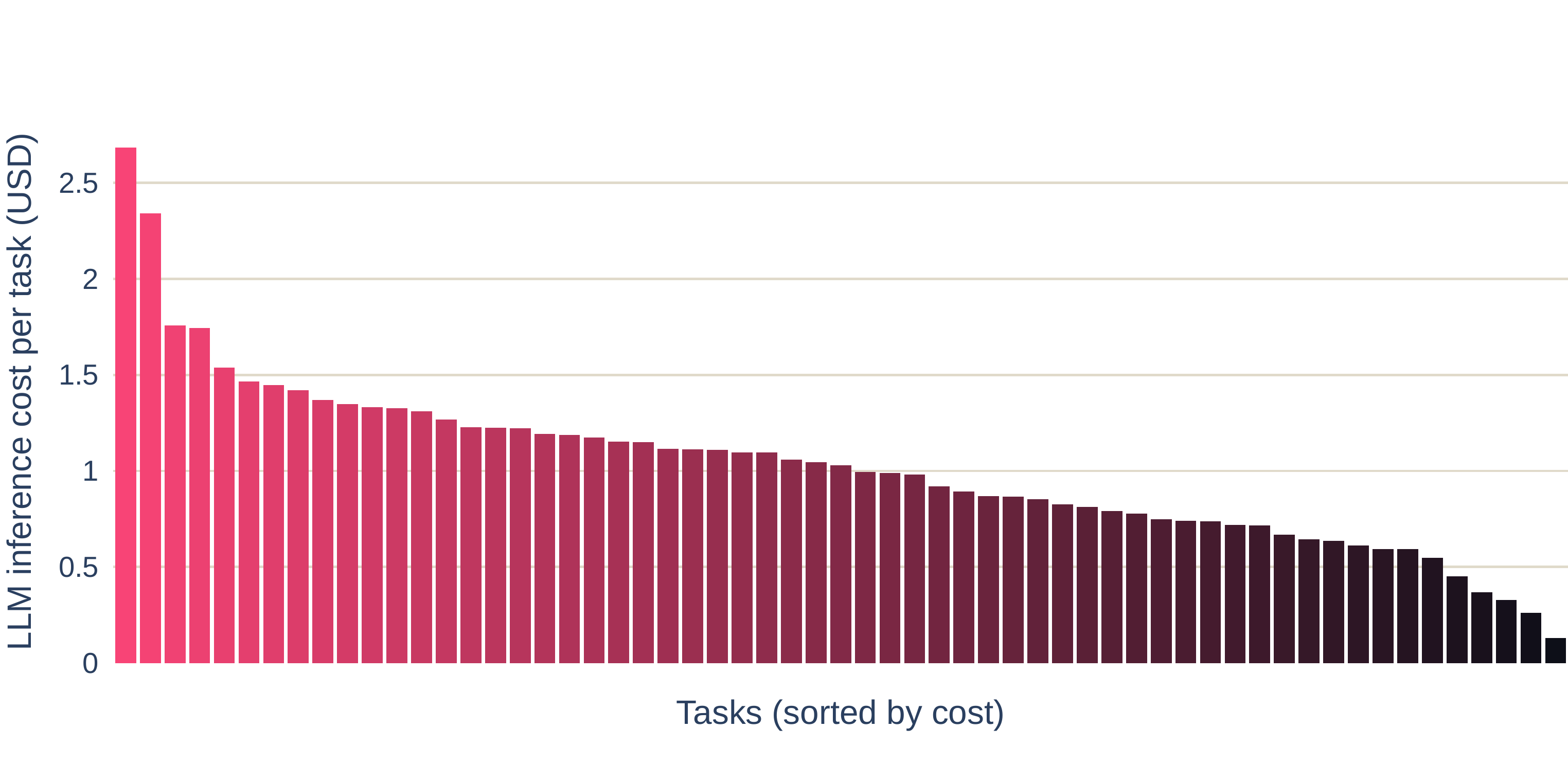}
    \caption{LLM cost of running AIDE per task on the Weco-Kaggle benchmark. We used GPT-4 Turbo (\texttt{gpt-4-0125-preview}) with pricing data from January to February 2024.}
    \label{fig:llm-cost-aide}
\end{figure}

Figure~\ref{fig:llm-cost-aide} illustrates the per-task LLM inference cost for AIDE across the Weco-Kaggle benchmark, using GPT-4 Turbo (\texttt{gpt-4-0125-preview}) with pricing data from early 2024. Although certain tasks incur higher costs due to more extensive prompting (up to approximately \$2.50 per task), the majority remain under \$1.50, reflecting moderate token usage and minimal manual intervention. Overall, these expenditures are much lower than the investment required for human experts or conventional AutoML services, especially when considering the significant performance gains achieved by AIDE’s fully automated design. Moreover, as language model costs continue to decline, AIDE’s approach becomes increasingly competitive in terms of both performance and budget. 





\section{Weco Kaggle Benchmark}
\label{appendix:weco-kaggle-full}

\footnotesize 
\setlength{\extrarowheight}{2pt} 

\begin{tabularx}{\linewidth}{l c c c l l}
\toprule
Competition & Local Eval & Kaggle Submittable & GPU & Data Size  \\
\midrule
\endfirsthead
\toprule
Competition & has local eval & kaggle submittable & needs gpu & compressed data size  \\
\midrule
\endhead
\href{https://www.kaggle.com/c/bike-sharing-demand}{bike-sharing-demand} & \checkmark & \checkmark &  & 193.8 kB & \\
\href{https://www.kaggle.com/c/cat-in-the-dat}{cat-in-the-dat} & \checkmark & \checkmark &  & 22.3 MB & \\
\href{https://www.kaggle.com/c/godaddy-microbusiness-density-forecasting}{godaddy-microbusiness-density-forecasting} & \checkmark &  &  & 1.9 MB & \\
\href{https://www.kaggle.com/c/house-prices-advanced-regression-techniques}{house-prices-advanced-regression-techniques} & \checkmark & \checkmark &  & 203.8 kB & \\
\href{https://www.kaggle.com/c/icr-identify-age-related-conditions}{icr-identify-age-related-conditions} & \checkmark & \textasciitilde &  & 154.1 kB & \\
\href{https://www.kaggle.com/c/new-york-city-taxi-fare-prediction}{new-york-city-taxi-fare-prediction} & \checkmark & \checkmark &  & 1.7 GB & \\
\href{https://www.kaggle.com/c/optiver-trading-at-the-close}{optiver-trading-at-the-close} & \checkmark &  &  & 210.3 MB & \\
\href{https://www.kaggle.com/c/playground-series-s3e14}{playground-series-s3e14} & \checkmark & \checkmark &  & 649.5 kB & \\
\href{https://www.kaggle.com/c/playground-series-s3e16}{playground-series-s3e16} & \checkmark & \checkmark &  & 2.8 MB & \\
\href{https://www.kaggle.com/c/playground-series-s3e17}{playground-series-s3e17} & \checkmark &  &  & 3.7 MB & \\
\href{https://www.kaggle.com/c/playground-series-s3e18}{playground-series-s3e18} & \checkmark &  &  & 2.5 MB & \\
\href{https://www.kaggle.com/c/playground-series-s3e19}{playground-series-s3e19} & \checkmark & \checkmark &  & 1.2 MB & \\
\href{https://www.kaggle.com/c/playground-series-s3e20}{playground-series-s3e20} & \checkmark &  &  & 51.3 MB & \\
\href{https://www.kaggle.com/c/playground-series-s3e22}{playground-series-s3e22} & \checkmark & \checkmark &  & 61.2 kB & \\
\href{https://www.kaggle.com/c/playground-series-s3e23}{playground-series-s3e23} & \checkmark &  &  & 6.0 MB & \\
\href{https://www.kaggle.com/c/playground-series-s3e24}{playground-series-s3e24} & \checkmark & \checkmark &  & 7.1 MB & \\
\href{https://www.kaggle.com/c/playground-series-s3e25}{playground-series-s3e25} & \checkmark & \checkmark &  & 575.1 kB & \\
\href{https://www.kaggle.com/c/tabular-playground-series-aug-2022}{tabular-playground-series-aug-2022} & \checkmark & \checkmark &  & 2.4 MB & \\
\href{https://www.kaggle.com/c/tabular-playground-series-feb-2021}{tabular-playground-series-feb-2021} & \checkmark & \checkmark &  & 68.9 MB & \\
\href{https://www.kaggle.com/c/tabular-playground-series-feb-2022}{tabular-playground-series-feb-2022} & \checkmark & \checkmark &  & 279.7 MB & \\
\href{https://www.kaggle.com/c/tabular-playground-series-jan-2022}{tabular-playground-series-jan-2022} & \checkmark & \checkmark &  & 236.0 kB & \\
\href{https://www.kaggle.com/c/tabular-playground-series-jul-2021}{tabular-playground-series-jul-2021} & \checkmark & \checkmark &  & 270.4 kB & \\
\href{https://www.kaggle.com/c/tmdb-box-office-prediction}{tmdb-box-office-prediction} & \checkmark & \checkmark &  & 18.3 MB & \\
\href{https://www.kaggle.com/c/career-con-2019}{career-con-2019} &  & \checkmark &  & 36.5 MB & \\
\href{https://www.kaggle.com/c/carvana-image-masking-challenge}{carvana-image-masking-challenge} &  & \checkmark & \checkmark & 26.2 GB & \\
\href{https://www.kaggle.com/c/cat-in-the-dat-ii}{cat-in-the-dat-ii} &  & \checkmark &  & 43.3 MB & \\
\href{https://www.kaggle.com/c/cifar-10}{cifar-10} &  & \checkmark & \checkmark & 750.1 MB & \\
\href{https://www.kaggle.com/c/ciphertext-challenge-ii}{ciphertext-challenge-ii} &  & \checkmark & \checkmark & 95.6 MB & \\
\href{https://www.kaggle.com/c/ciphertext-challenge-iii}{ciphertext-challenge-iii} &  & \checkmark & \checkmark & 34.4 MB & \\
\href{https://www.kaggle.com/c/competitive-data-science-predict-future-sales}{competitive-data-science-predict-future-sales} &  & \checkmark &  & 15.8 MB & \\
\href{https://www.kaggle.com/c/digit-recognizer}{digit-recognizer} &  & \checkmark &  & 16.1 MB & \\
\href{https://www.kaggle.com/c/dog-breed-identification}{dog-breed-identification} &  & \checkmark & \checkmark & 724.5 MB & \\
\href{https://www.kaggle.com/c/dont-overfit-ii}{dont-overfit-ii} &  & \checkmark & \checkmark & 13.3 MB & \\
\href{https://www.kaggle.com/c/facial-keypoints-detection}{facial-keypoints-detection} &  & \checkmark & \checkmark & 80.0 MB & \\
\href{https://www.kaggle.com/c/forest-cover-type-prediction}{forest-cover-type-prediction} &  & \checkmark & \checkmark & 26.6 MB & \\
\href{https://www.kaggle.com/c/histopathologic-cancer-detection}{histopathologic-cancer-detection} &  & \checkmark & \checkmark & 6.8 GB & \\
\href{https://www.kaggle.com/c/home-data-for-ml-course}{home-data-for-ml-course} &  & \checkmark &  & 395.3 kB & \\
\href{https://www.kaggle.com/c/iwildcam-2019-fgvc6}{iwildcam-2019-fgvc6} &  & \checkmark & \checkmark & 46.6 GB & \\
\href{https://www.kaggle.com/c/jigsaw-toxic-comment-classification-challenge}{jigsaw-toxic-comment-classification-challenge} &  & \checkmark & \checkmark & 55.2 MB & \\
\href{https://www.kaggle.com/c/kuzushiji-recognition}{kuzushiji-recognition} &  & \checkmark & \checkmark & 4.5 GB & \\
\href{https://www.kaggle.com/c/nlp-getting-started}{nlp-getting-started} &  & \checkmark &  & 607.3 kB & \\
\href{https://www.kaggle.com/c/noaa-right-whale-recognition}{noaa-right-whale-recognition} &  & \checkmark & \checkmark & 9.8 GB & \\
\href{https://www.kaggle.com/c/planttraits2024}{planttraits2024} &  & \checkmark & \checkmark & 3.8 GB & \\
\href{https://www.kaggle.com/c/playground-series-s3e1}{playground-series-s3e1} &  & \checkmark &  & 2.4 MB & \\
\href{https://www.kaggle.com/c/playground-series-s3e11}{playground-series-s3e11} &  & \checkmark &  & 9.4 MB & \\
\href{https://www.kaggle.com/c/playground-series-s3e13}{playground-series-s3e13} &  & \checkmark &  & 21.0 kB & \\
\href{https://www.kaggle.com/c/playground-series-s3e15}{playground-series-s3e15} &  & \checkmark &  & 414.8 kB & \\
\href{https://www.kaggle.com/c/playground-series-s3e26}{playground-series-s3e26} &  & \checkmark &  & 358.8 kB & \\
\href{https://www.kaggle.com/c/playground-series-s3e3}{playground-series-s3e3} &  & \checkmark &  & 97.4 kB & \\
\href{https://www.kaggle.com/c/playground-series-s3e5}{playground-series-s3e5} &  & \checkmark &  & 69.6 kB & \\
\href{https://www.kaggle.com/c/playground-series-s3e7}{playground-series-s3e7} &  & \checkmark &  & 929.8 kB & \\
\href{https://www.kaggle.com/c/playground-series-s3e9}{playground-series-s3e9} &  & \checkmark &  & 110.9 kB & \\
\href{https://www.kaggle.com/c/playground-series-s4e1}{playground-series-s4e1} &  & \checkmark &  & 7.1 MB & \\
\href{https://www.kaggle.com/c/playground-series-s4e2}{playground-series-s4e2} &  & \checkmark &  & 939.5 kB & \\
\href{https://www.kaggle.com/c/predict-volcanic-eruptions-ingv-oe}{predict-volcanic-eruptions-ingv-oe} &  & \checkmark & \checkmark & 10.2 GB & \\
\href{https://www.kaggle.com/c/recognizing-faces-in-the-wild}{recognizing-faces-in-the-wild} &  & \checkmark & \checkmark & 399.8 MB & \\
\href{https://www.kaggle.com/c/rsna-pneumonia-detection-challenge}{rsna-pneumonia-detection-challenge} &  & \checkmark & \checkmark & 3.9 GB & \\
\href{https://www.kaggle.com/c/santa-2019-revenge-of-the-accountants}{santa-2019-revenge-of-the-accountants} &  & \checkmark &  & 95.7 kB & \\
\href{https://www.kaggle.com/c/scrabble-player-rating}{scrabble-player-rating} &  & \checkmark &  & 39.2 MB & \\
\href{https://www.kaggle.com/c/sentiment-analysis-on-movie-reviews}{sentiment-analysis-on-movie-reviews} &  & \checkmark &  & 2.0 MB & \\
\href{https://www.kaggle.com/c/spaceship-titanic}{spaceship-titanic} &  & \checkmark &  & 306.4 kB & \\
\href{https://www.kaggle.com/c/tabular-playground-series-apr-2021}{tabular-playground-series-apr-2021} &  & \checkmark &  & 4.6 MB & \\
\href{https://www.kaggle.com/c/tabular-playground-series-apr-2022}{tabular-playground-series-apr-2022} &  & \checkmark &  & 179.6 MB & \\
\href{https://www.kaggle.com/c/tabular-playground-series-aug-2021}{tabular-playground-series-aug-2021} &  & \checkmark &  & 156.7 MB & \\
\href{https://www.kaggle.com/c/tabular-playground-series-dec-2021}{tabular-playground-series-dec-2021} &  & \checkmark &  & 131.8 MB & \\
\href{https://www.kaggle.com/c/tabular-playground-series-jan-2021}{tabular-playground-series-jan-2021} &  & \checkmark &  & 65.8 MB & \\
\href{https://www.kaggle.com/c/tabular-playground-series-jul-2022}{tabular-playground-series-jul-2022} &  & \checkmark &  & 20.6 MB & \\
\href{https://www.kaggle.com/c/tabular-playground-series-jun-2021}{tabular-playground-series-jun-2021} &  & \checkmark &  & 10.4 MB & \\
\href{https://www.kaggle.com/c/tabular-playground-series-jun-2022}{tabular-playground-series-jun-2022} &  & \checkmark &  & 245.9 MB & \\
\href{https://www.kaggle.com/c/tabular-playground-series-mar-2021}{tabular-playground-series-mar-2021} &  & \checkmark &  & 57.6 MB & \\
\href{https://www.kaggle.com/c/tabular-playground-series-mar-2022}{tabular-playground-series-mar-2022} &  & \checkmark &  & 4.9 MB & \\
\href{https://www.kaggle.com/c/tabular-playground-series-may-2021}{tabular-playground-series-may-2021} &  & \checkmark &  & 2.8 MB & \\
\href{https://www.kaggle.com/c/tabular-playground-series-may-2022}{tabular-playground-series-may-2022} &  & \checkmark &  & 269.9 MB & \\
\href{https://www.kaggle.com/c/tabular-playground-series-nov-2021}{tabular-playground-series-nov-2021} &  & \checkmark &  & 449.0 MB & \\
\href{https://www.kaggle.com/c/tabular-playground-series-nov-2022}{tabular-playground-series-nov-2022} &  & \checkmark &  & 1.1 GB & \\
\href{https://www.kaggle.com/c/tabular-playground-series-oct-2021}{tabular-playground-series-oct-2021} &  & \checkmark &  & 1.4 GB & \\
\href{https://www.kaggle.com/c/tabular-playground-series-oct-2022}{tabular-playground-series-oct-2022} &  & \checkmark &  & 3.7 GB & \\
\href{https://www.kaggle.com/c/tabular-playground-series-sep-2021}{tabular-playground-series-sep-2021} &  & \checkmark &  & 626.2 MB & \\
\href{https://www.kaggle.com/c/tabular-playground-series-sep-2022}{tabular-playground-series-sep-2022} &  & \checkmark &  & 630.1 kB & \\
\href{https://www.kaggle.com/c/tgs-salt-identification-challenge}{tgs-salt-identification-challenge} &  & \checkmark & \checkmark & 466.1 MB & \\
\bottomrule
\end{tabularx}

\end{document}